\documentclass[10pt]{article}

\usepackage[letterpaper, textwidth=5.5in, textheight=9in, top=1in, headheight=12pt,
            headsep=25pt, footskip=30pt]{geometry}
\usepackage{latexsym}
\usepackage{booktabs}
\usepackage{graphicx}
\usepackage{multirow}
\usepackage{amsmath}
\usepackage{microtype}
\usepackage{inconsolata}
\usepackage{enumitem}
\usepackage{wrapfig}
\usepackage{float}
\usepackage[font=small, labelfont=bf, skip=5pt]{caption}
\usepackage{titlesec}
\usepackage{xcolor}
\usepackage{tikz}
\usepackage{pgfplots}
\pgfplotsset{compat=1.18}
\usetikzlibrary{positioning,arrows.meta,fit,backgrounds,calc,shapes.geometric,decorations.pathreplacing,patterns}
\usepackage[numbers, square, sort&compress]{natbib}
\usepackage{url}
\definecolor{darkblue}{rgb}{0, 0, 0.5}
\usepackage[colorlinks=true, citecolor=darkblue, linkcolor=darkblue,
            urlcolor=darkblue]{hyperref}

\usepackage{fontspec}
\usepackage{fontawesome5}
\newfontfamily\deva{NewCM10Devanagari-Regular.otf}[Script=Devanagari]
\newcommand{\np}[1]{{\deva #1}}
\newcommand{\nps}[1]{{\deva\small #1}}

\titleformat{\section}{\large\bfseries}{\thesection}{1em}{}
\titleformat{\subsection}{\normalsize\bfseries}{\thesubsection}{1em}{}
\titlespacing*{\section}{0pt}{12pt}{6pt}
\titlespacing*{\subsection}{0pt}{9pt}{4pt}
\titlespacing*{\paragraph}{0pt}{5.5pt}{0.6em}
\setlist{nosep, leftmargin=1.6em}
\makeatletter
\def\ps@firstpage{\def\@oddhead{}\def\@evenhead{}%
  \def\@oddfoot{\small Preprint.\hfil}\def\@evenfoot{\small Preprint.\hfil}}
\renewcommand{\@maketitle}{%
  \vbox{\hsize\textwidth\centering
    \hrule height 4pt\vskip 0.25in
    {\LARGE\bfseries \@title\par}
    \vskip 0.22in\hrule height 1pt\vskip 0.3in
    {\normalsize \@author\par}}
  \vskip 0.3in}
\renewenvironment{abstract}{%
  \centerline{\large\bfseries Abstract}\vspace{0.5ex}%
  \list{}{\leftmargin=0.5in\rightmargin=0.5in}\item\relax}{\endlist\vspace{0.5ex}}
\makeatother

\newcommand{\papaya}{Papaya}
\newcommand{\cut}{\textbar}
\newcommand{\sd}[1]{{\scriptsize$\pm$#1}}

\newcommand\blfootnote[1]{%
  \begingroup
  \renewcommand\thefootnote{}\footnote{#1}%
  \addtocounter{footnote}{-1}%
  \endgroup
}

\title{Preserving Morphemes: Morphology-Guided Pre-Tokenization for Nepali}

\author{%
  \begin{tabular}[t]{c}
    \textbf{Kalash Shrestha}\\ Kathmandu University\\ \texttt{kalashestha@gmail.com}
  \end{tabular}\hspace{0.5in}%
  \begin{tabular}[t]{c}
    \textbf{Nikhil Pradhan}\\ IOE Thapathali Campus\\ \texttt{nikhilpradhan20b@gmail.com}
  \end{tabular}\\[0.8em]
  \raisebox{-0.1em}{\large\faGithub}\hspace{0.6em}%
  \href{https://github.com/kalexrt/nepali-morphology-tokenizer}{\texttt{https://github.com/kalexrt/nepali-morphology-tokenizer}}}
\date{}

\begin{document}
\maketitle
\thispagestyle{firstpage}

\begin{abstract}
A byte-level BPE vocabulary learns each inflected form of a Nepali word as a separate
string, so the stem \np{सरकार} `government' is spelled three different ways in
\np{सरकारले}, \np{सरकारको} and \np{सरकारमा}. We test whether splitting words into stem
and affixes before BPE helps, with the corpus, vocabulary size, model and number of
training steps held fixed. Our pre-tokenizer, \papaya{}, uses a finite-state transducer
built from a published grammar of Nepali, falls back to regular expressions, and leaves the
BPE trainer unchanged. On 607 words annotated by seven native speakers its segmenter
reaches 0.96 boundary F1, and the resulting tokens keep stems intact far more often than
plain BPE does. In a 17M-parameter language model it lowers bits per byte by about 1\,\%
at equal training steps; most of the larger gain seen at equal epochs comes from the extra
steps that longer token sequences buy, and an unsupervised Morfessor segmentation gives
the same improvement. Downstream the effect is small: NER improves only on entities that
contain words unseen in training, POS tagging and news classification do not change, and
published Nepali tokenizers perform about as well. We release the annotated boundary set,
a 556-affix dataset and the code.
\end{abstract}

\blfootnote{Code, the affix dataset, the boundary set and the trained tokenizers are
available at \url{https://github.com/kalexrt/nepali-morphology-tokenizer}, the data under CC BY 4.0 and the code under the MIT licence. The
affix tables are our re-derivation of \citet{prasain2011}; the author's own transducers
are not redistributed. Devanagari copied from this PDF may come out with pre-base vowel
signs reordered, an artefact of the PDF text layer; the released files are the
reference.}

\section{Introduction}\label{sec:intro}

\begin{figure}[t]
\centering
\begin{minipage}[c]{0.46\textwidth}
\resizebox{\linewidth}{!}{%
\begin{tikzpicture}[
  font=\small, node distance=0pt,
  stage/.style={draw, rounded corners=1pt, minimum height=5mm, minimum width=5.4cm,
                align=center, inner ysep=2.5pt},
  ours/.style={stage, fill=black!12, very thick},
  stock/.style={stage, fill=white, densely dashed},
  ar/.style={-{Latex[length=1.6mm]}, shorten >=1pt, shorten <=1pt},
  tag/.style={font=\scriptsize\itshape, anchor=west},
  whole/.style={draw, minimum height=4.6mm, inner xsep=2.5pt, fill=black!6},
  stem/.style={draw, very thick, minimum height=4.6mm, inner xsep=2.5pt, fill=black!20},
  clit/.style={draw, minimum height=4.6mm, inner xsep=2.5pt, fill=white},
  lab/.style={font=\scriptsize\itshape, anchor=east}
]
% the mechanism: one entry per form vs one shared stem
\node[lab] at (-2.05,1.95) {plain BPE};
\node[whole, anchor=west] (w1) at (-1.95,1.95) {\nps{सरकारले}};
\node[whole, right=1.5mm of w1] (w2) {\nps{सरकारको}};
\node[whole, right=1.5mm of w2] (w3) {\nps{सरकारमा}};
\node[lab] at (-2.05,1.3) {\papaya};
\node[stem, anchor=west] (s1) at (-1.95,1.3) {\nps{सरकार}};
\node[clit, right=0pt of s1] (c1) {\nps{ले}};
\node[stem, right=1.5mm of c1] (s2) {\nps{सरकार}};
\node[clit, right=0pt of s2] (c2) {\nps{को}};
\node[stem, right=1.5mm of c2] (s3) {\nps{सरकार}};
\node[clit, right=0pt of s3] {\nps{मा}};
\draw[black!40] (-2.8,0.72) -- (2.8,0.72);

% the pipeline
\node[stage, fill=white] (raw) at (0,0.25) {raw text\ \ \nps{सरकारले गरेको}};
\node[ours, below=3.2mm of raw] (fst) {FST tier over an induced stem lexicon};
\node[ours, below=2.2mm of fst] (rx) {regex tier, for words the FST leaves};
\node[stage, fill=white, below=3.2mm of rx] (mark) {marked text\ \ \nps{सरकार}\cut\nps{ले} \nps{गरे}\cut\nps{को}};
\node[stock, below=3.2mm of mark] (split) {\texttt{Split(U+2060, removed)}};
\node[stock, below=2.2mm of split] (byte) {\texttt{ByteLevel} + BPE merges};
\node[stage, fill=white, below=3.2mm of byte] (tok) {token ids};

\draw[ar] (raw) -- (fst);
\draw[ar] (rx) -- (mark);
\draw[ar] (mark) -- (split);
\draw[ar] (byte) -- (tok);

\draw[decorate, decoration={brace, amplitude=3pt}, black!60]
  ($(fst.north east)+(0.08,0)$) -- ($(rx.south east)+(0.08,0)$)
  node[tag, midway, xshift=3pt] {ours};
\draw[decorate, decoration={brace, amplitude=3pt}, black!60]
  ($(split.north east)+(0.08,0)$) -- ($(byte.south east)+(0.08,0)$)
  node[tag, midway, xshift=3pt] {stock};
\end{tikzpicture}}
\end{minipage}\hfill
\begin{minipage}[c]{0.49\textwidth}
\caption{Top: plain BPE spends one vocabulary entry on each frequent inflected form;
\papaya{} cuts the clitics off first, so every form shares one stem entry. Below: the
pipeline. Only the two shaded stages are ours; they insert U+2060 \textsc{word joiner}
at each cut, and the unmodified \texttt{tokenizers} pipeline splits on it before any
merge. Decoding reproduces the input byte for byte.}
\label{fig:pipeline}
\end{minipage}
\end{figure}
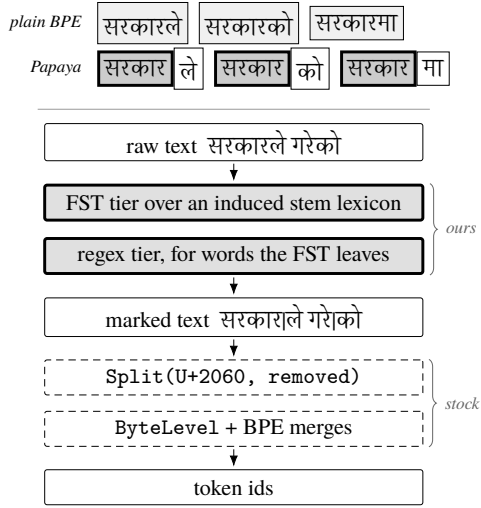

Nepali marks number and case on the noun with clitics and tense and agreement on the verb
with suffixes, so one stem surfaces in many forms: \np{सरकार} `government' also appears as
\np{सरकारले}, \np{सरकारको} and \np{सरकारमा}. A frequency-driven subword vocabulary sees
four frequent strings and spends four entries on them, so the model learns the stem's
distribution once per form rather than once (Figure~\ref{fig:pipeline}). A form of a
known stem that did not make the merge table is then re-segmented arbitrarily.

Nepali is a good place to test whether repairing this helps. It is morphologically rich
and low-resource, and it also has a complete published computational grammar
\citep{prasain2011} whose paradigm tables cover the inflectional system. Prior
morphology-aware tokenizers segment with a statistical model before BPE
\citep{luitel2025perplexity, taboasgarcia2025exploring}, constrain BPE's merges from
inside \citep{asgari2026morphbpe, brahma2025morphtok}, or replace BPE with a lexicon
lookup \citep{bayram2025tokens}. None takes its segmentation from a descriptive grammar
of Nepali.

We ask a narrower question than ``does our tokenizer win'': whether aligning tokens to
morphemes buys anything measurable on a matched corpus, vocabulary and model, once the
obvious confounds are controlled. That framing answers \citet{vemula2025rethinking}, who
find that morphological alignment does not predict downstream performance, more usefully
than a leaderboard result would. Our contributions are three.

\begin{enumerate}[nosep]
\item \textbf{\papaya}, a rule-based pre-tokenizer (a finite-state transducer over a
corpus-induced stem lexicon, with a regex fallback) composed with unmodified byte-level
BPE. The segmentation is preprocessing, so the trainer and decoder are stock library code.
\item Two resources: a 607-word Nepali morpheme-boundary set, adjudicated by seven native
speakers, and a machine-readable dataset of 556 affixes re-derived from
\citet{prasain2011}.
\item A step- and seed-matched evaluation against plain BPE, Unigram, Morfessor-based
segmentation and published Nepali tokenizers, in which a step-count control accounts for
two thirds of the apparent language-modelling gain.
\end{enumerate}

\papaya{} aligns with annotated boundaries far better than plain BPE and lowers bits per
byte by 1.1\,\% at equal optimizer steps. Once pretraining steps are matched, its NER gain is significant
only on unseen entities, and fragmenting every word does as well there. POS tagging and topic classification do not move, and
against published Nepali tokenizers the result is parity.

\section{Related Work}\label{sec:related}

\paragraph{Morphology-aware tokenization.}
One line of work segments before BPE and leaves the algorithm alone:
\citet{luitel2025perplexity} pre-segment Nepali with Morfessor
\citep{virpioja2013morfessor}, \citet{taboasgarcia2025exploring} use a semi-supervised
segmenter for Spanish. A second line changes BPE itself: MorphBPE \citep{asgari2026morphbpe} forbids
merges across gold morpheme boundaries, and MorphTok \citep{brahma2025morphtok} constrains
BPE for Hindi and Marathi. A third replaces BPE for known words with a root-and-affix
lookup \citep{bayram2025tokens}. \papaya{} belongs to the first line, but takes its
segmentation from a descriptive grammar rather than from a model fitted to the corpus.

\paragraph{Nepali.}
Nepali subword work has so far been about the script rather than the morphology.
\citet{regmi2026vowel} show that the standard letter-only pre-tokenization regex severs
Devanagari vowel signs from their base character and train a Devanagari-aware BPE on the fix. We adopt the fix as our
baseline and keep the unfixed regex as a control, so that we measure morphology on top of
a correct script treatment. NepBERTa \citep{timilsina2022nepberta} and the Nepali BPE of
\citet{shrestha2025towards} use standard vocabularies. On the
resource side, \citet{prasain2011} gives a full computational description of Nepali
inflection, and \texttt{apertium-ne} \citep{forcada2011apertium}
provides an independent affix list; neither has been connected to subword tokenization.

\paragraph{Does alignment matter?}
\citet{vemula2025rethinking} find, for Telugu, that the choice between Unigram and BPE
matters more than morphological alignment, and that intrinsic alignment does not predict
downstream performance. \citet{arnett2025alignment} evaluate alignment across 70
languages. We test Vemula et al.'s claim directly, with Unigram and
Unigram-over-\papaya{} rows, on a language whose morphology is unusually well described,
and two of our three downstream tasks return the null result it predicts.

\section{Methodology}\label{sec:method}

\papaya{} cuts a Nepali word into a stem and its affixes before byte-level BPE sees it
(Figure~\ref{fig:pipeline}). A tier scans each word, a run of Devanagari letters and
combining marks, and rejoins the pieces it licenses with U+2060 \textsc{word joiner}, a
zero-width character absent from the corpus. The tokenizer's first pre-tokenizer stage
splits on the marker and deletes it, so no merge can span a proposed cut. The trainer,
merge rules, vocabulary file and decoder are stock library code; we change the input
string, not the algorithm, so porting the idea to another language means replacing the
tier and nothing else. Two tiers exist: a lexicon-free regex tier, which is our ablation
arm, and an FST tier, which falls back to the regex tier on words it cannot analyse. The
FST tier with that fallback is \papaya{}.

Both tiers draw on an affix dataset that we re-derived, in our own schema, from the rule
tables of \citet{prasain2011}, Chapters~3--6: 660 rows, 556 unique nominal, verbal,
indeclinable and derivational affixes, each with provenance, a confidence value and a
\texttt{regex}/\texttt{fst} tier recording whether attaching it leaves the stem's spelling
unchanged. The dissertation's Devanagari text layer is corrupted by its font encoding; we
repaired it by OCR anchored on the intact IPA column, with 233 rule
rows confirmed by hand (Appendix~\ref{app:ocr}).

\subsection{The regex tier}\label{sec:method:regex}

The regex tier peels suffixes off the right edge of a word by longest match. It admits
the affixes that our affix dataset marks as concatenating without changing
the stem's spelling, plus the genitive \np{को}/\np{का}/\np{की} and the emphatic case forms
(\np{मै}, \np{बाटै}), which the dataset assigns to the FST tier because their \emph{form}
is not derivable by concatenation even though the \emph{cut} is clean. The \texttt{core}
profile holds 20 affixes (the plural \np{हरू}, the case clitics and two emphatic
particles); \texttt{full} adds 133 postpositional forms and 11 verbal suffixes, for 164
(Appendix~\ref{app:affix}). Cuts obey three rules: the stem keeps at least one full
letter; a nominal cut never follows a halanta, so the surname \np{शर्मा}, whose \np{र्म} is
one conjunct, is not cut before \np{मा}; and a verbal cut only follows a halanta and never
splits a geminate. An ordering grammar, [stem]\,[\np{हरू}]\,[postp $\mid$
case]$^{*}$\,[\textsc{gen} $\mid$ verbal] with at most four cuts, is applied right to left
(examples in Appendix~\ref{app:affix}). Without a lexicon the tier cannot tell stems from
look-alikes: vowel-final verb stems such as \np{खानु} stay whole, and \np{ढोका} `door' is
cut.

\subsection{The FST tier}\label{sec:method:fst}

\begin{figure}[b]
\centering
\begin{minipage}[c]{0.55\textwidth}
\resizebox{\linewidth}{!}{%
\begin{tikzpicture}[
  font=\scriptsize,
  n/.style={draw, rounded corners=1pt, minimum height=4mm, inner xsep=2.5pt, align=center},
  lane/.style={n, fill=black!8},
  cls/.style={n, fill=white},
  ar/.style={-{Latex[length=1.3mm]}},
  xar/.style={ar, densely dashed, black!55}
]
\node[n, fill=black!18] (root) at (0,-1.2) {Root};
\node[cls] (pre) at (1.25,-1.2) {Prefix\\\nps{न}-};

\node[lane] (vc) at (3.0,-0.15) {VC\ \nps{गर्}};
\node[lane] (vv) at (3.0,-0.75) {VV\ \nps{खा}};
\node[lane] (vu) at (3.0,-1.35) {VU\ \nps{पठाउ}};
\node[lane] (ns) at (3.0,-2.3) {NStem\ \nps{केटा}};

\node[cls] (vsuf) at (5.4,-0.75) {verbal suffix\\\nps{नु ने एको छ यो दा}};
\node[cls] (nsuf) at (5.4,-2.3) {\nps{हरू} $\rightarrow$ case $\rightarrow$\\postp $\rightarrow$ \textsc{gen}};

\draw[ar] (root) -- (pre);
\draw[ar] (pre.east) -- (vc.west);
\draw[ar] (pre.east) -- (vv.west);
\draw[ar] (pre.east) -- (vu.west);
\draw[ar] (pre.east) -- (ns.west);
\draw[ar] (vc.east) -- (vsuf.west);
\draw[ar] (vv.east) -- (vsuf.west);
\draw[ar] (vu.east) -- (vsuf.west);
\draw[ar] (ns.east) -- (nsuf.west);

\draw[xar] (vsuf.south) to[out=-70, in=20] (nsuf.north east);
\draw[xar] (vc.south west) to[out=-120, in=130] (ns.north west);

\node[font=\scriptsize\itshape, anchor=west, text width=3.3cm] at (-0.4,-3.2)
  {dashed: a verbal participle declines like a noun; a verb stem may also be
   a nominal one};
\end{tikzpicture}}
\end{minipage}\hfill
\begin{minipage}[c]{0.41\textwidth}
\caption{The segmenting transducer, abridged. Verb stems come in three shapes,
consonant-final (VC), vowel-final (VV) and polysyllabic -\nps{उ} (VU), because the
stem/suffix join is spelled differently in each. The crossover arcs are why a ranking step
is needed: \nps{भर्ना} has both a verbal and a nominal parse.}
\label{fig:transducer}
\end{minipage}
\end{figure}
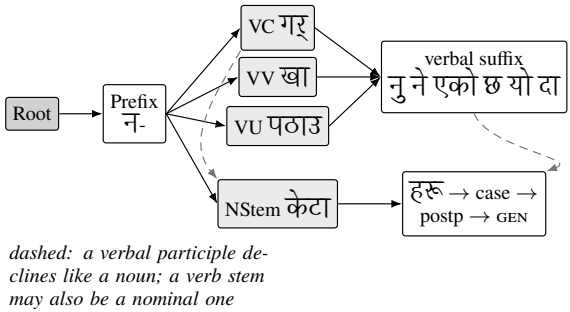

The FST tier cuts a word only where a known stem and a licensed chain of surface
allomorphs spell it exactly. It is a \texttt{lexc} grammar compiled with foma
\citep{hulden2009foma}: the lower side is the surface word, the upper side the same string
with a boundary symbol at each cut, so a lookup returns every licensed segmentation.
Sandhi is expressed as allomorph choice in the continuation classes, with separate lanes
for the three verb-stem shapes and for nominals (Figure~\ref{fig:transducer};
Appendix~\ref{app:fst}).

The lexicon is induced from word-type counts in the tokenizer-training corpus and seeded
from the dissertation's annex word lists. Verb stems need several attested inflected
slots that nouns never fill and an attested infinitive; nominal stems need standalone
attestation and several case markers, with stricter thresholds for short strings; and
frequent, pluralised ``strong'' nouns outrank a verbal parse of the same string, so
\np{खाना} `food' is not \np{खा}+\np{ना}. Induction yields 720 verbal and 18{,}568 nominal
stems (thresholds in Appendix~\ref{app:fst}). The transducer then audits its own
lexicon: we compile once, drop every nominal candidate it analyses as an inflected form of
a more frequent, bound or verbal stem (\np{उनीहरू} over \np{उनी}), and recompile.

A word may receive several analyses. We prefer one that is verbal with a cut or headed by
a strong noun, then the longest stem (\np{भूमिका} over \np{भूमि}+\np{का}), then the fewest
cuts. An assertion checks that the pieces concatenate back to the input, so the tier only
inserts boundaries. In \texttt{lex} mode an unanalysable word stays whole; in
\texttt{hybrid} mode, which is \papaya{}, it goes to the regex tier's \texttt{full}
profile unless it occurs at least 100 times in the corpus, since a frequent unanalysable
word is usually a lexicalised look-alike such as \np{निकै}.

Cuts fall at akshara boundaries only, because a token may never begin with a combining
mark: underlying \np{गर्}+\np{एको} surfaces as \np{गरे}\cut\np{को}, and \np{ठिकै} stays
whole. The gold set is normalised to the same convention before scoring
(\S\ref{sec:data}).

\subsection{Composing with BPE}\label{sec:method:bpe}

The marked text enters a standard byte-level BPE whose pre-tokenizer is
\texttt{Split(U+2060, removed)}, then a GPT-2-style word split that keeps Devanagari
combining marks with their consonant, then the byte-level mapping, so every cut is a hard
pre-token boundary and the vocabulary has no unknown token. Because the tier never edits a
character and the marker never enters the token stream, decoding returns the original
bytes, which we assert at build time.

\subsection{Design choices}\label{sec:method:choices}

We run the FST tier first, in \texttt{hybrid} mode, because that cascade was the most
accurate segmenter we built. On the gold set the regex tier alone scores 0.800 boundary
F1, the FST tier without a fallback 0.946 and with it 0.961; running the regex tier first
drops to 0.904 even with a precision-tuned regex (Table~\ref{tab:ablation}). Two later
refinements, a homograph list and a stricter fallback, raise tier F1 to 0.969 but change
neither bits per byte nor NER (Table~\ref{tab:ablation}), so we report the first version,
the only one with the full vocabulary sweep and the step-matched control. The language
model is a small GPT-2-style decoder (Table~\ref{tab:hyper}), small enough that every
variant could be pretrained with three seeds on one GPU; 16{,}000 is the main vocabulary
size, and the 8{,}000 and 32{,}000 runs check that the conclusions do not depend on it.

\subsection{Data and gold set}\label{sec:data}

All systems are trained and evaluated on Nepali web text from FineWeb-2
\citep{penedo2025fineweb2}, subset \texttt{npi\_Deva} (ODC-By 1.0). A 200{,}000-document
file supplies the word counts for stem induction, and its first 100{,}000 documents fit
every tokenizer; a separate 400{,}000-document file is the language-model pretraining data,
seen once by every variant; the first 5{,}000 documents of the test split (1.69M words)
serve all intrinsic and language-modelling evaluation. Every variant sees identical text,
and nothing is trained on the test split.

For boundary evaluation we sampled 610 word types from the test split in six buckets that
cover both the constructions the tiers handle and the look-alikes they must leave alone.
Seven native speakers, unpaid volunteers including both authors, marked every
inflectional and clitic boundary (\np{केटा}\cut\np{हरू}\cut\np{लाई}) by adjudicated
consensus, and three unresolved words were removed, leaving 607 (Appendix~\ref{app:annot}).
Annotators consulted one another, so the set is a consensus reference and we report no
agreement coefficient. The regex baseline's segmentation was shown beside 422 randomly
chosen words and withheld on 185; on the blind rows the regex tier scores 0.773 F1 (0.812
shown) and \papaya{}'s tier 0.980 (0.953 shown), so the suggestion did not favour \papaya{}.
Because the buckets follow what the tiers do and no separate development split exists,
boundary F1 on this set is not an estimate for running text. Two silver sets, from
UD\_Nepali-BK and a Hunspell affix file, give an outside check (Appendix~\ref{app:ocr}).

\section{Experimental Setup}\label{sec:setup}

\paragraph{Tokenizers.}
Every in-house tokenizer is stock byte-level BPE \citep{sennrich2016neural,
radford2019language} from HuggingFace \texttt{tokenizers} \citep{moi2023tokenizers} at a
16{,}000 vocabulary; they differ only in what runs in front of it. \emph{Plain BPE}, the
baseline, splits on \texttt{[\textbackslash p\{L\}\textbackslash p\{M\}]+}, keeping vowel
signs with their consonant. The \emph{mark-split control} uses the GPT-2 letter class,
which severs every vowel sign from its akshara \citep{regmi2026vowel}. \emph{Regex full}
runs the regex tier alone. \emph{Morfessor + BPE} marks the cuts of a Morfessor Baseline
model \citep{virpioja2013morfessor} trained on the same word counts over akshara units.
\emph{Unigram} replaces BPE with a Unigram LM on plain BPE's pre-tokenizer, and
\emph{\papaya{} + Unigram} runs Unigram on \papaya{}'s marked text. Plain BPE and
\papaya{} are also trained at 8{,}000 and 32{,}000. We score ten published tokenizers
intrinsically (Appendix~\ref{app:ext}), and put NepBERTa \citep{timilsina2022nepberta},
Sakonii \citep{sakonii2022distilbert}, Llama~3.1 \citep{grattafiori2024llama3} and
Qwen~2.5 \citep{qwen2024qwen25} through our language model and fine-tuning in place of our
vocabulary. These four are not parameter-matched to ours (23M, 20M, 60M and 69M parameters
against 17M, almost all in the embedding table).

\paragraph{Language model and the matched-steps control.}
We pretrain one decoder-only Transformer per tokenizer (6 layers, $d = 384$, context 512;
17M parameters at 16{,}000) for one epoch with three seeds (Appendix~\ref{app:hyper}). A
tokenizer with higher fertility takes more optimizer steps per epoch: \papaya{} takes
3{,}061 steps at 16{,}000 to plain BPE's 2{,}633 ($+16.3\,\%$). We therefore also train
plain BPE and Morfessor + BPE for \papaya{}'s step
count, setting the epoch count to the fertility ratio (to within 0.2\,\% of the steps).
The control favours the baseline, which sees up to a fifth of its documents twice, and we
fine-tune from the matched plain-BPE checkpoints too.

\paragraph{Downstream tasks.}
Named entity recognition (3{,}289 sentences, 7 BIO labels) and part-of-speech tagging
(39 tags) come from Nep-gLUE \citep{timilsina2022nepberta}, split 80/10/10. Content
classification (CC) uses the NepaliNewsClassification set from MTEB \citep{mteb2026nepalinews}:
3 classes, 2{,}048 training and 1{,}495 test paragraphs of running text. We fine-tune the whole model with a linear head
with 5 fine-tuning seeds spread over the 3 pretraining checkpoints (two each on the
first two, one on the third). Nep-gLUE text is pre-split at word level,
with clitics already standing as separate words, so the tier has less to do there than on
running text.

\paragraph{Metrics.}
\emph{Fertility} is tokens per whitespace word on the test documents. \emph{Boundary F1} compares annotated cut positions with where a segmenter's pieces or a
tokenizer's tokens begin. \emph{Consistency} is the F1 between the tokens inside a stem's
span in an inflected form and the tokens of the bare stem, over the 344 inflected gold
forms. \emph{Bits per byte} (bpb) is test negative log-likelihood divided by UTF-8 bytes,
comparable across vocabularies. NER is scored by conlleval span F1, POS and CC by macro F1. Comparisons use a one-sided paired
bootstrap over test sentences \citep{koehn2004bootstrap} with 2{,}000 resamples of the
seed-averaged difference.

\section{Results}\label{sec:results}

\begin{table}[t]
\centering\footnotesize
\setlength{\tabcolsep}{2.6pt}
\begin{tabular}{@{}lcccccccc@{}}
\toprule
& & & & & \multicolumn{2}{c}{NER} & & \\
\cmidrule(lr){6-7}
Tokenizer & Tok./w & Bound. & Cons. & bpb & all & unseen & POS & CC \\
\midrule
Mark-split control     & 4.40 & 0.177 & 0.902 & 0.5124$^\dagger$ & 0.765\sd{.010} & 0.600 & 0.926\sd{.004} & 0.964\sd{.006} \\
Plain BPE              & 1.40 & 0.218 & 0.305 & 0.4971 & 0.736\sd{.009} & 0.508 & 0.929\sd{.003} & 0.972\sd{.004} \\
\quad at \papaya{}'s step count & --- & --- & --- & 0.4863 & 0.755\sd{.011} & 0.529 & 0.929\sd{.003} & 0.974\sd{.005} \\
Unigram                & 1.78 & 0.525 & 0.613 & 0.4849 & 0.758\sd{.005} & 0.585 & 0.929\sd{.004} & 0.971\sd{.003} \\
Morfessor + BPE        & 1.46 & 0.457 & 0.443 & 0.4871 & 0.747\sd{.005} & 0.567 & 0.930\sd{.002} & 0.974\sd{.002} \\
Regex full + BPE       & 1.58 & 0.709 & 0.823 & 0.4820 & 0.761\sd{.015} & 0.598 & 0.930\sd{.002} & 0.974\sd{.002} \\
\textbf{\papaya{}}     & 1.63 & \textbf{0.859} & 0.903 & \textbf{0.4809} & 0.764\sd{.006} & 0.592 & 0.930\sd{.002} & 0.975\sd{.004} \\
\papaya{} + Unigram    & 1.91 & 0.783 & \textbf{0.936} & 0.4807$^\S$ & 0.756\sd{.009} & 0.588 & 0.928\sd{.001} & 0.974\sd{.003} \\
\midrule
NepBERTa (30.5k)       & \textbf{1.28} & 0.266 & 0.290 & 0.4866 & 0.762\sd{.010} & 0.581 & 0.928\sd{.005} & \textbf{0.978}\sd{.004} \\
Sakonii (24.6k)        & 1.34 & 0.448 & 0.417 & 0.4867 & 0.757\sd{.007} & 0.588 & \textbf{0.932}\sd{.003} & 0.973\sd{.004} \\
Llama 3.1 (128k)$^\ddagger$ & 3.57 & 0.119 & 0.795 & 0.5059 & 0.758 & \textbf{0.609} & 0.923 & 0.962 \\
Qwen 2.5 (152k)$^\ddagger$  & 6.19 & 0.222 & 0.913 & 0.5169 & \textbf{0.769} & 0.586 & 0.922 & 0.956 \\
\bottomrule
\end{tabular}
\caption{All systems at a 16{,}000 vocabulary unless noted, on identical text. Boundary F1
is scored on the 607-word gold set; bold marks the best value in each column, for bpb
among rows trained for no more steps than \papaya{}; consistency is stem-token F1 on its 344 inflected
forms, and must be read next to fertility, since a near-character-level split is
consistent by construction (the mark-split control, Qwen~2.5). bpb is at one epoch, three
seeds; the indented row is plain BPE trained for \papaya{}'s step count. NER is span F1,
POS and CC macro F1, mean \sd{sd} over five fine-tuning seeds; ``unseen'' is NER span F1
on the 117 entities per seed that contain a word absent from the NER training data.
$^\dagger$8{,}315 steps, not step-matched. $^\S$3{,}603 steps; 0.4897 at \papaya{}'s step count. $^\ddagger$One seed; not parameter-matched
(\S\ref{sec:setup}). Regex core and gated rows and six further published tokenizers are in
Appendix~\ref{app:ext}.}
\label{tab:main}
\end{table}

\subsection{Intrinsic evaluation}\label{sec:results:intrinsic}

As segmenters, before BPE, \papaya{}'s tier scores 0.961 boundary F1 on the gold set
(precision 0.959, recall 0.962), against 0.800 for the regex tier and 0.466 for Morfessor,
which rarely cuts (recall 0.306). Through BPE (Table~\ref{tab:main}) \papaya{} scores 0.859
against plain BPE's 0.218, a comparison close to true by construction, since \papaya{}
inserts the boundaries it is scored on; recall stays at 0.969 while precision falls to
0.771, because at a 16{,}000 vocabulary BPE re-cuts inside stems. The advantage reproduces
on both silver sets (0.789 against 0.277 F1 on the UD sample, 0.896 against 0.660 recall
on the Hunspell set). \papaya{} also keeps stems stable across inflection, 0.903
consistency against 0.305, but the mark-split control ties it at 0.902 by cutting every
vowel sign off, at 4.40 tokens per word against \papaya{}'s 1.63 and plain BPE's 1.40.

\subsection{Language modelling}\label{sec:results:lm}

At equal optimizer steps \papaya{} lowers bits per byte by 0.9\,\%, 1.1\,\% and 1.6\,\% at
8{,}000, 16{,}000 and 32{,}000 (Table~\ref{tab:bpb}); at equal epochs the gaps read 2.2\,\%,
3.3\,\% and 4.7\,\%, so about two thirds of the one-epoch gap is the extra optimizer steps
that higher fertility buys. Plain BPE is flat in vocabulary while \papaya{} improves at
every size. The regex tier is within about one seed standard deviation of \papaya{} (0.4820
against 0.4809), and Morfessor + BPE reaches 0.4802 at \papaya{}'s step count: an
unsupervised segmenter matches \papaya{} on bits per byte, so the language-modelling gain
over plain BPE does not need the grammar.

\begin{table}[t]
\centering
\begin{minipage}[c]{0.5\textwidth}
\centering\small
\setlength{\tabcolsep}{3.5pt}
\begin{tabular}{@{}lccccc@{}}
\toprule
& \multicolumn{2}{c}{Plain BPE} & & \multicolumn{2}{c}{$\Delta$ \papaya} \\
\cmidrule(lr){2-3}\cmidrule(lr){5-6}
Vocab & 1 ep. & matched & \papaya{} & eq. steps & eq. ep. \\
\midrule
8{,}000  & 0.4987 & 0.4924 & \textbf{0.4879} & $-0.9\,\%$ & $-2.2\,\%$ \\
16{,}000 & 0.4971 & 0.4863 & \textbf{0.4809} & $-1.1\,\%$ & $-3.3\,\%$ \\
32{,}000 & 0.4970 & 0.4811 & \textbf{0.4735} & $-1.6\,\%$ & $-4.7\,\%$ \\
\bottomrule
\end{tabular}
\end{minipage}\hfill
\begin{minipage}[c]{0.46\textwidth}
\caption{Bits per byte, three seeds per cell; bold is best per row. ``Matched'' is plain BPE trained for
\papaya{}'s step count; we report the equal-steps difference, and show the equal-epoch one
beside it because the gap between them is optimizer steps.}
\label{tab:bpb}
\end{minipage}
\end{table}

\subsection{Downstream tasks}\label{sec:results:downstream}

From one-epoch checkpoints \papaya{} gains $+0.028$ NER span F1 over plain BPE (95\,\%
interval $[+0.014, +0.041]$, $p < 0.001$), but \papaya{} had 16\,\% more pretraining
steps. Against plain BPE at \papaya{}'s step count the gain shrinks to $+0.009$
($[-0.003, +0.022]$, $p = 0.07$) and is not significant. POS ($+0.001$, $p = 0.34$) and
CC ($+0.000$, $p = 0.50$) do not move.

\paragraph{Where the gain comes from.}
On the entities that contain a word absent from the NER training data, \papaya{} raises
span F1 from 0.508 to 0.592, and this part survives the step control: against plain BPE at
\papaya{}'s step count (0.529) the difference is $+0.063$ ($[+0.017, +0.114]$,
$p = 0.001$). On entities made of frequent words the gain is 1.5 points. Fragmentation
reaches the same effect: the mark-split control scores 0.600 on unseen entities and ties
\papaya{} on NER overall ($-0.001$, $p = 0.56$), at 2.7 times the sequence length and the
worst bits per byte in Table~\ref{tab:main}. The gap is constant, $+0.02$ to $+0.03$, from
10\,\% to 100\,\% of the NER training data (Appendix~\ref{app:ablation}), so it is not a
data-efficiency effect.

\paragraph{Published tokenizers.}
Every published tokenizer scores at most 0.448 boundary F1 (Sakonii). Against NepBERTa and
Sakonii the downstream result is parity: \papaya{} $-$ NepBERTa on NER is $+0.002$
($p = 0.40$) and \papaya{} $-$ Sakonii $+0.006$ ($p = 0.20$). Both reach 0.4866--0.4867
bpb, better than plain BPE and worse than \papaya{}. Llama~3.1 and Qwen~2.5 are worse
language models than our baseline, at 3.57 and 6.19 tokens per word.

\paragraph{Tier variants.}
Four transducer-based variants spanning 0.904--0.969 tier F1 land within 0.759--0.769 NER
and 0.4805--0.4809 bpb (Appendix~\ref{app:ablation}); at this scale the downstream tasks
saturate below the segmentation quality the tiers reach, so tier changes must be judged on
boundary F1 and consistency.

\section{Conclusion}\label{sec:conclusion}

Tokens that follow Nepali morphemes align with human boundaries and keep stems stable far
better than plain BPE, at a cost of 16\,\% more tokens per word, and lower bits per byte by
0.9--1.6\,\% once optimizer steps are matched. An unsupervised Morfessor segmentation matches
that bits-per-byte gain, so what the grammar adds is boundary accuracy. Downstream, the
benefit that survives the step control is confined to NER on unseen entities, which bounds
rather than overturns the null result of \citet{vemula2025rethinking}: alignment matters
where the model must generalise to unseen surface forms, a condition that averaged
benchmarks hide. Fragmentation reaches the same unseen-entity result, but at a higher cost
in sequence length and language-model loss; the morphological route gets there at
near-baseline length. Against published Nepali tokenizers the result is parity.

\paragraph{Future work.}
All language models here have 17M parameters; whether the equal-steps bits-per-byte gap
survives at 50--100M parameters is the question that decides whether grammar-guided
pre-tokenization is worth adopting at scale. On the tier side, recall is limited by verbs
too rare to clear the induction thresholds (\np{नसड्ने}) and by compound verbs whose first
element is not a listed form (\np{देखापर्न}); homographs such as \np{भर्ना} `admission',
cut after the verb \np{भर्} `fill', are a ranking error that more data cannot fix.
Derivational cuts can be added once an evaluation set marks them, and a blind
re-annotation would give the gold set an inter-annotator agreement measure. Finally, because the
tier is preprocessing, porting it to other morphologically rich South Asian languages with
a published grammar, or using it to extend the vocabulary of a larger multilingual LLM,
requires only a new tier; neither has been tested here.

\section*{Use of Generative AI}

We used generative AI assistants (Claude, via Claude Code) to speed up engineering and
writing: to develop the code of the segmentation, training and evaluation pipeline, to run
and organise experiments, and to draft and edit the text of this paper. All boundary annotations were produced by the native-speaker annotators, and
no AI-generated data was used as gold labels. The authors verified every reported number
against the result files and take full responsibility for the content.

\clearpage
\bibliographystyle{acl_natbib}
\bibliography{refs}

\clearpage
\appendix
\begin{center}{\LARGE\bfseries Appendix}\end{center}
\vspace{0.5em}
\addcontentsline{toc}{section}{Appendix}

\section{Gold set protocol}\label{app:annot}

Annotators marked every inflectional and clitic cut (case markers, postpositions, plural,
emphatics, participle and infinitive endings, bound pronoun stems); in compound verbs only
suffixes are cut, so \np{गर्नुपर्ने} is \np{गर्नु}\cut\np{पर्}\cut\np{ने}. The regex tier's
segmentation was shown beside a random 70\,\% of words and withheld on 30\,\%. The 610
sampled types fall into six buckets (natural 250, verbal 120, gate-blocked 80,
postpositional 60, emphatic 40, unsplit 60). A word was settled once two or three
annotators agreed, harder words went to all seven, and three that stayed ambiguous
(\nps{सैफले}, \nps{लड्नका}, \nps{मैले}) were removed; 17 annotations were moved to
akshara boundaries, with the raw form kept. All annotators consented to the release of
their annotations.

\section{Affix dataset and silver sets}\label{app:ocr}

The dataset's 660 rows come from Chapters~3--6 of \citet{prasain2011} (120 nominal, 247
verbal, 172 indeclinable, 121 derivational); the tier field is \texttt{regex} if attaching
the affix leaves the stem's spelling unchanged and \texttt{fst} otherwise (286 and 270 of
the 556 unique affixes; three of the \texttt{fst} rows are zero-derivation, flagged as
not affixation and unused by the tiers). Rows were cross-checked against the author's XFST source and
\texttt{apertium-ne} \citep{forcada2011apertium}. Because the PDF's font
encoding corrupts its Devanagari text layer, an IPA-anchored check flagged 963 of 2{,}027
annex entries; OCR of the rendered pages with Chandra OCR~2\footnote{\url{https://github.com/datalab-to/chandra}}, aligned on the
intact IPA and gloss columns, recovered 80.8\,\% of them, and 233 rule rows and 87 annex
entries were confirmed by hand.

The UD silver set holds the 85 tokens of UD\_Nepali-BK v2.18 \citep{bk2026udnepali} whose
lemma is a proper prefix of the surface form. The Hunspell set holds 3{,}223 synthetic
form/lemma pairs from the \texttt{ne\_NP} affix file via NepaliLemmatizer\footnote{\url{https://github.com/dpakpdl/NepaliLemmatizer}}, marking only the stem boundary, so we report recall only.
FineWeb-2 is ODC-By 1.0, UD\_Nepali-BK and NepaliNewsClassification CC BY-SA 4.0,
NepaliLemmatizer MIT, and Nep-gLUE is used under its research terms.

\section{Segmentation tiers}\label{app:affix}\label{app:fst}

\paragraph{Regex tier.} \texttt{core}: \np{हरू}, \np{ले}, \np{लाई}, \np{को}, \np{का},
\np{की}, \np{कै}, \np{मा}, \np{मै}, \np{बाट}, \np{बाटै}, \np{सँग}, \np{सँगै}, \np{सित},
\np{सितै}, \np{तिर}, \np{तिरै}, \np{देखि}, \np{ही}, \np{सुकै}. \texttt{full} adds 133
postpositional forms from Chapter~5 and 11 verbal suffixes (\np{नु}, \np{ने}, \np{ना},
\ldots) for 164. Right to left, nothing is cut left of \np{हरू}, the genitive may follow
any marker, a postposition may follow a case marker, and a non-genitive case marker never
follows another. The attestation gate (stem seen $\geq 100$ times standalone) moves tier
precision from 0.790 to 0.906 and recall from 0.811 to 0.689.

\paragraph{FST tier.} A verb stem needs at least 4 inflected slots attested at least 3
times each, at least 3 of them slots nouns never fill (\np{छ}, \np{यो}, \np{दा},
\np{एको}, \ldots), and an attested infinitive. A nominal stem needs standalone occurrence
and at least 2 case markers, with minima raised to 100 and 3 for one- or two-character
stems and 20 and 3 for three. A strong noun (at least 8 markers, 1{,}000 standalone
occurrences, attested with \np{हरू}) outranks a verbal parse; 1{,}108 stems are strong.
Consonant stems (\np{गर्}) cut after the halanta; vowel stems (\np{दि}, \np{हु}) take
\np{न्छ} in the non-past and a chandrabindu before \np{द}-initial endings, and
polysyllabic \np{-उ} stems (\np{बनाउ}) drop the \np{उ} before \np{यो}, \np{ए} and \np{इ}.
Postpositions missing from Chapter~5 are a declared extra group: \np{भन्दा}, \np{बिना},
\np{द्वारा}, \np{लगायत}, \np{सम्बन्धी}, \np{झैँ}.

\section{Further intrinsic results}\label{app:ext}

Table~\ref{tab:ext} lists the systems left out of Table~\ref{tab:main}. Regex core cuts
only 20 affixes, so its boundary F1 stays at 0.494. The attestation gate raises the regex
tier's own precision but, through BPE, leaves boundary F1 (0.700) and consistency (0.782)
slightly below regex full, and neither variant moves bits per byte or NER. None of the six
further published tokenizers exceeds 0.43 boundary F1. The high consistency of mBERT and
Gemma~2 comes with 2.8--2.9 tokens per word, while NepaliBPE, the most compact at 1.24,
keeps stems intact least often (0.061).

\begin{table}[H]
\centering\small
\setlength{\tabcolsep}{3pt}
\begin{tabular}{@{}lrcccc@{}}
\toprule
Tokenizer & Vocab & Tok/w & Bound. & Cons. & bpb / NER \\
\midrule
Regex core    & 16{,}000 & 1.55 & 0.494 & 0.510 & 0.4839 / 0.759 \\
Regex + gate  & 16{,}000 & 1.56 & \textbf{0.700} & 0.782 & \textbf{0.4836} / \textbf{0.762} \\
NepaliBPE$^a$ & 50{,}006 & \textbf{1.24} & 0.197 & 0.061 & --- \\
arkios$^b$    & 65{,}536 & 1.42 & 0.225 & 0.317 & --- \\
mBERT$^c$     & 119{,}547 & 2.82 & 0.248 & 0.846 & --- \\
gpt-4o$^d$    & 200{,}000 & 2.11 & 0.248 & 0.642 & --- \\
XLM-R$^e$     & 250{,}002 & 1.67 & 0.428 & 0.622 & --- \\
Gemma 2$^f$   & 256{,}000 & 2.89 & 0.360 & \textbf{0.905} & --- \\
\bottomrule
\end{tabular}
\caption{Rows not in Table~\ref{tab:main}, same measures; bold is best per column. Hub ids:
$^a$\nolinkurl{Aananda-giri/NepaliBPE},
$^b$\nolinkurl{sajalregmi4/arkios-tokenizer} \citep{regmi2026vowel},
$^c$\nolinkurl{bert-base-multilingual-cased}, $^d$\nolinkurl{Xenova/gpt-4o},
$^e$\nolinkurl{xlm-roberta-base}, $^f$\nolinkurl{unsloth/gemma-2-2b}.}
\label{tab:ext}
\end{table}

\section{Tier variants and NER data curve}\label{app:ablation}

Table~\ref{tab:ablation} is the ablation behind the design choices in
\S\ref{sec:method:choices}; Table~\ref{tab:curve} repeats NER fine-tuning on 10--100\,\%
of the training sentences.

\begin{table}[H]
\centering\small
\setlength{\tabcolsep}{1.8pt}
\begin{tabular}{@{}lcccccc@{}}
\toprule
& \multicolumn{2}{c}{Bound.\ F1} & & & \multicolumn{2}{c}{NER} \\
\cmidrule(lr){2-3}\cmidrule(l){6-7}
Variant & tier & BPE & Cons. & bpb & all & unseen \\
\midrule
\papaya{}          & 0.961 & 0.859 & \textbf{0.903} & 0.4809 & 0.764 & 0.592 \\
+ homographs & 0.965 & \textbf{0.863} & \textbf{0.903} & \textbf{0.4805} & 0.766 & \textbf{0.606} \\
\ + strict fallb. & \textbf{0.969} & \textbf{0.863} & \textbf{0.903} & 0.4807 & 0.759 & 0.580 \\
\ \ regex first & 0.904 & 0.804 & 0.879 & 0.4808 & \textbf{0.769} & 0.594 \\
\bottomrule
\end{tabular}
\caption{Tier variants at 16k, each adding to the row above: a list of 19 lexicalised
homographs; a regex fallback without 19 low-precision affixes; the cascade reversed. Bold is best per column. No NER
pair differs from \papaya{} at $p < 0.05$.}
\label{tab:ablation}
\end{table}

\begin{table}[H]
\centering\small
\setlength{\tabcolsep}{3pt}
\begin{tabular}{@{}lcccc@{}}
\toprule
NER training data & 10\,\% & 25\,\% & 50\,\% & 100\,\% \\
\midrule
Plain BPE   & 0.653 & 0.686 & 0.714 & 0.736 \\
Regex full  & \textbf{0.678} & 0.706 & 0.727 & 0.761 \\
\papaya{}   & 0.677 & \textbf{0.715} & \textbf{0.735} & \textbf{0.764} \\
\bottomrule
\end{tabular}
\caption{NER span F1 by share of the NER training sentences, five seeds; bold is best per column.}
\label{tab:curve}
\end{table}

\section{Hyperparameters}\label{app:hyper}

\begin{table}[H]
\centering\small
\setlength{\tabcolsep}{3pt}
\begin{tabular}{@{}ll@{}}
\toprule
\multicolumn{2}{@{}l}{\emph{Pretraining}} \\
architecture & 6 layers, 6 heads, $d=384$, pre-LN, GELU \\
context & 512, learned positions, tied embeddings \\
params (8k/16k/32k) & 14M / 17M / 23M \\
optimizer & AdamW $(0.9, 0.95)$, wd 0.1, clip 1.0 \\
learning rate & $1.2\times10^{-3}$, 200 warm-up, cosine \\
batch & 128 $\times$ 512 tokens, bf16 \\
data / seeds & 400{,}000 documents, 1 epoch / 1--3 \\
\midrule
\multicolumn{2}{@{}l}{\emph{Fine-tuning}} \\
learning rate & $1\times10^{-4}$, 5\,\% warm-up, cosine \\
batch / epochs & 16 / 8, best development epoch \\
seeds & 1--5 \\
\bottomrule
\end{tabular}
\caption{Training configuration. The matched-steps control sets the epoch count to the
fertility ratio (1.116, 1.164, 1.210 at 8k/16k/32k). Pretraining took about 29 GPU-hours on a single GPU,
reruns included.}
\label{tab:hyper}
\end{table}

\end{document}